\documentclass[a4paper, 10pt, conference]{ieeeconf}      

\IEEEoverridecommandlockouts                              

\usepackage{graphicx} 
\usepackage{color}
\usepackage{flushend}

\title{\LARGE \bf
LGSVL Simulator: A High Fidelity Simulator for Autonomous Driving
}

\author{Guodong Rong$^{1}$, Byung Hyun Shin$^{1}$, Hadi Tabatabaee$^{1}$, Qiang Lu$^{1}$, Steve Lemke$^{1}$, M\={a}rti\c{n}\v{s} Mo\v{z}eiko$^{1}$, \\
Eric Boise$^{1}$, Geehoon Uhm$^{1}$, Mark Gerow$^{1}$, Shalin Mehta$^{1}$, Eugene Agafonov$^{1}$, Tae Hyung Kim$^{1}$, \\
Eric Sterner$^{1}$, Keunhae Ushiroda$^{1}$, Michael Reyes$^{1}$, Dmitry Zelenkovsky$^{1}$, Seonman Kim$^{1}$
\thanks{$^{1}$LG Electronics America R\&D Lab. {Corresponding authors: \tt\small\{dmitry.zelenkovsky, seonman.kim\}@lge.com}}
}

\begin{document}

\maketitle
\thispagestyle{empty}
\pagestyle{empty}

\begin{abstract}
Testing autonomous driving algorithms on real autonomous vehicles is extremely costly and many researchers and developers in the field cannot afford a real car and the corresponding sensors. Although several free and open-source autonomous driving stacks, such as Autoware and Apollo are available, choices of open-source simulators to use with them are limited. In this paper, we introduce the \emph{LGSVL Simulator} which is a high fidelity simulator for autonomous driving. The simulator engine provides end-to-end, full-stack simulation which is ready to be hooked up to Autoware and Apollo. In addition, simulator tools are provided with the core simulation engine which allow users to easily customize sensors, create new types of controllable objects, replace some modules in the core simulator, and create digital twins of particular environments.

\end{abstract}

\section{INTRODUCTION}
\label{Sec_Intruduction}

Autonomous vehicles have seen dramatic progress in the past several years. Research shows that autonomous vehicles have to be driven billions of miles to demonstrate their liability \cite{KS2016}, which is impossible without the help of simulation. From the very beginning of autonomous driving research \cite{Pomerleau1989}, simulators have played a key role in development and testing of autonomous driving (AD) stacks. Simulation allows developers to quickly test new algorithms without driving real vehicles. Compared to road testing, simulation has several important advantages: It is safer than real road testing, particularly for some dangerous scenarios (e.g. pedestrian jaywalking), and can generate corner cases which are rarely encountered in the real world (e.g. extreme weather). Moreover, a simulator is able to exactly reproduce all factors of a problematic scenario and thus allows developers to debug and test new patches.

Today's autonomous driving systems utilize deep neural networks (DNN) in more and more modules to help improve performance. Training DNN models requires a large amount of labeled data. Traditional datasets for autonomous driving, such as KITTI \cite{GLSU2013} and Cityscapes \cite{COR+2016}, do not have enough data for DNN to deal with complicated scenarios. Although several large datasets have been recently published by academia \cite{YXC+2018} and autonomous driving companies \cite{Nuscenes2019, Level5Dataset, WaymoOpenDataset}, these datasets which are collected from real world drives are usually manually (often with help from some automated tools) labeled, which is slow, costly, and error-prone. For some ground truth types, such as pixel-wise segmentation or optical flow, it is extremely difficult or impossible to manually label the data. Simulators can easily generate accurately labeled datasets that are an order-of-magnitude larger in size in parallel with the help of cloud platform.

\begin{figure}[tb]
\centering
    \includegraphics[width=.9\linewidth]{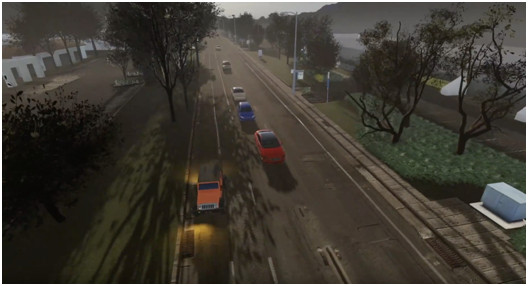}\\
    \includegraphics[width=.9\linewidth]{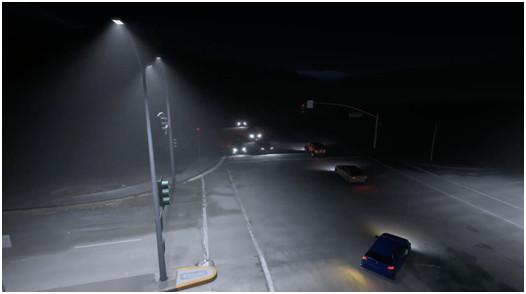}
\caption{Rendering examples by LGSVL Simulator}
\label{Fig_RenderingSamples}
\end{figure}

In this paper, we introduce the LGSVL Simulator\footnote{https://www.lgsvlsimulator.com/. ``LGSVL'' stands for ``LG Silicon Valley Lab'' which is now renamed to LG Electronics America R\&D Lab.}. The core simulation engine is developed using the Unity game engine \cite{Unity} and is open source with the source code freely available on GitHub\footnote{https://github.com/lgsvl/simulator}. The simulator has a communication bridge that enables passing messages between the simulator and an AD stack. By default the bridge supports ROS, ROS2, and Cyber RT messages, making it ready to be used with Autoware (ROS-based) and Baidu Apollo (ROS-based for 3.0 and previous versions, Cyber RT-based for 3.5 and later versions), the two most popular open source AD stacks. Map tools are provided to import and export HD maps for autonomous driving in formats such as Lanelet2 \cite{lanelet2}, OpenDRIVE, and Apollo HD Map. Fig.~\ref{Fig_RenderingSamples} illustrates some rendering examples from LGSVL Simulator.

The rest of this paper is organized as follows: Section \ref{Sec_RelatedWork} reviews prior related work. A detailed overview of the LGSVL Simulator is provided in Section \ref{Sec_OverviewOfLgsvlSimulator}. Some examples of applications of the simulator are listed in Section \ref{Sec_Applications}, and Section \ref{Sec_Conclusions} concludes the paper with direction of our future work.

\section{RELATED WORK}
\label{Sec_RelatedWork}

Simulation has been widely used in the automotive industry, especially for vehicle dynamics. Some famous examples are: CarMaker \cite{CarMaker}, CarSim \cite{CarSim}, and ADAMS \cite{ADAMS}. Autonomous driving requires more than just vehicle dynamics, and factors such as complex environment settings, different sensor arrangements and configurations, and simulating traffic for vehicles and pedestrians, must also be considered. Some of the earlier simulators \cite{TORCS, CSKX2015} run autonomous vehicles in virtual environments, but lack important features such as support for different sensors and simulating pedestrians.

Gazebo \cite{KH2004} is one of the most popular simulation platforms used in robotics and related research areas. Its modular design allows different sensor models and physics engines to be plugged into the simulator. But it is difficult to create large and complex environments with Gazebo and it lacks the newest advancements in rendering available in modern game engines like Unreal \cite{Unreal} and Unity.

There are some other popular open source simulators for autonomous driving, such as AirSim \cite{SDLK2017}, CARLA \cite{DRC+2017}, and Deepdrive \cite{Deepdrive}. These other simulators were typically created as research platforms to support reinforcement learning or synthetic data generation for machine learning, and usually require significant additional effort to integrate with a user's AD stack and communication bridge.

There are also several commercial automotive simulators including ANSYS \cite{ANSYS}, dSPACE \cite{dSPACE}, PreScan \cite{PreScan}, rFpro \cite{rFpro}, Cognata \cite{Cognata}, Metamoto \cite{Metamoto} and NVIDIA's Drive Constellation \cite{DriveConstellation}. However, because these simulators are not open source they can be difficult for users to customize to satisfy their own specific requirements or research goals.

Commercial video games related to driving nowadays offer realistic environments. Researchers have used games such as Grand Theft Auto V to generate synthetic datasets \cite{RVRK2016, RHK2017, RBM+2017}. However, this usually requires some hacking to be able to access resources in the game and can violate user license agreements. In addition, it is difficult if not impossible to support sensors other than a camera, and to deterministically control the vehicle as well as non-player characters such as pedestrians and traffic.

\section{OVERVIEW OF LGSVL SIMULATOR}
\label{Sec_OverviewOfLgsvlSimulator}

\begin{figure}[tb]
\centering
    \includegraphics[width=\linewidth]{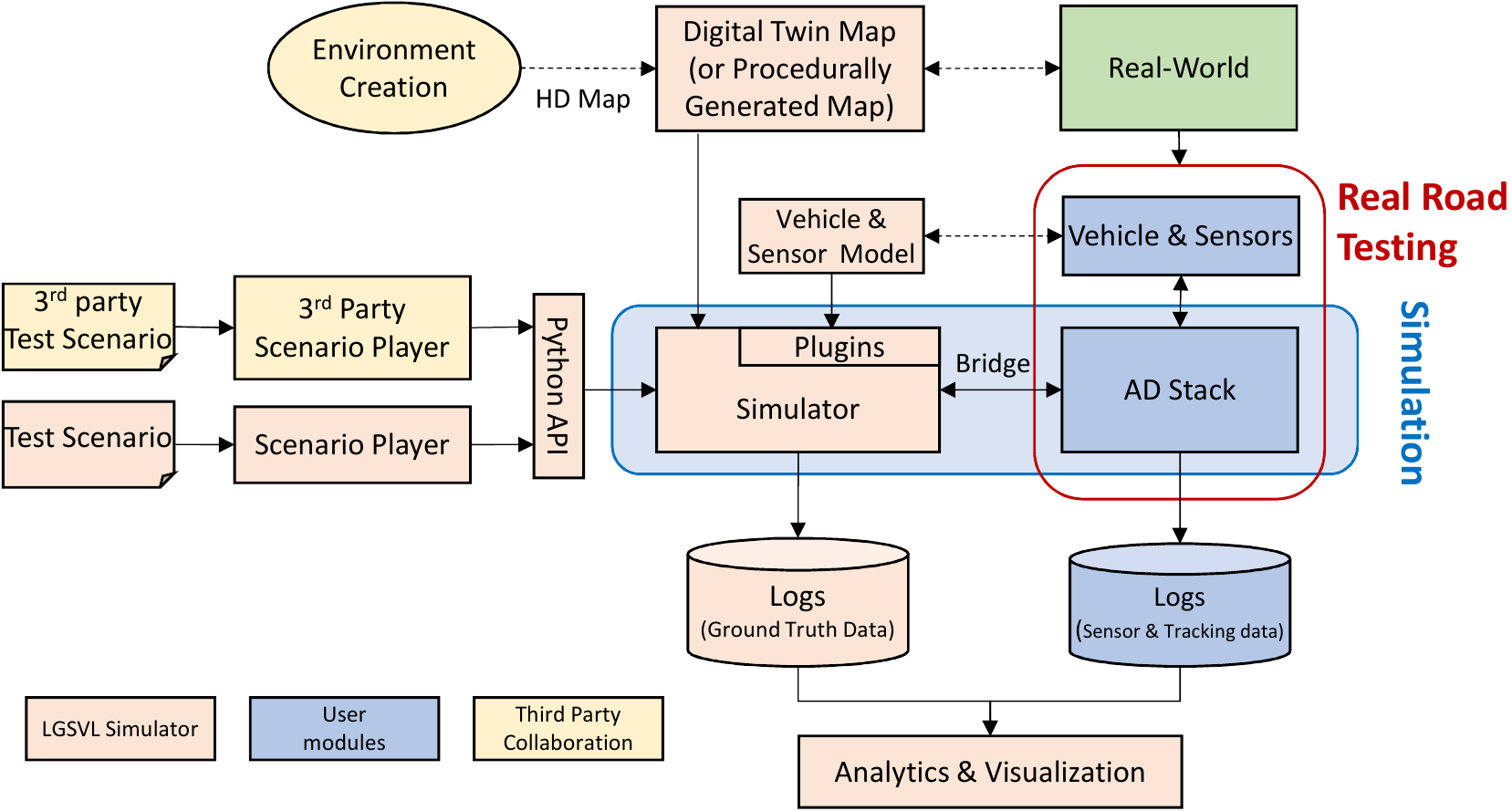}
\caption{Workflow of LGSVL Simulator}
\label{Fig_Workflow}
\end{figure}

The autonomous driving simulation workflow enabled by LGSVL Simulator is illustrated in Fig.~\ref{Fig_Workflow}. Details of each component are explained in the following of this section.

\subsection{User AD Stack}

The user AD stack is the system that the user wants to develop, test, and verify through simulation. LGSVL Simulator currently provides reference out-of-the-box integration with the open source AD system platforms Apollo\footnote{http://apollo.auto/}, developed by Baidu, and Autoware.AI\footnote{https://www.autoware.ai/} and Autoware.Auto\footnote{https://www.autoware.auto/}, developed by the Autoware Foundation. 

\begin{figure}[tb]
\centering
    \includegraphics[width=.9\linewidth]{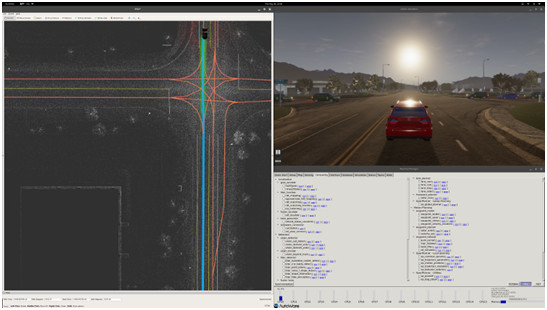}\\
    \includegraphics[width=.9\linewidth]{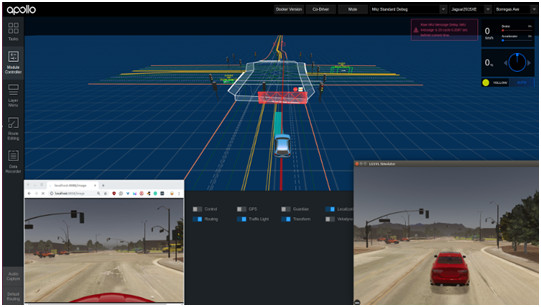}
\caption{Autoware (top) and Apollo (bottom) running with LGSVL Simulator}
\label{Fig_ADStacks}
\end{figure}

The user AD stack connects to LGSVL Simulator through a communication bridge interface; a bridge is selected based on the user AD stack's runtime framework. For Baidu's Apollo platform, which uses a custom runtime framework called Cyber RT, a custom bridge is provided to the simulator. Autoware.AI and Autoware.Auto, which run on ROS and ROS2, can connect to LGSVL Simulator through standard open source ROS and ROS2 bridges. Fig.~\ref{Fig_ADStacks} shows Autoware and Apollo running with LGSVL Simulator.

\begin{figure*}[tb]
\centering
    \includegraphics[width=.94\linewidth]{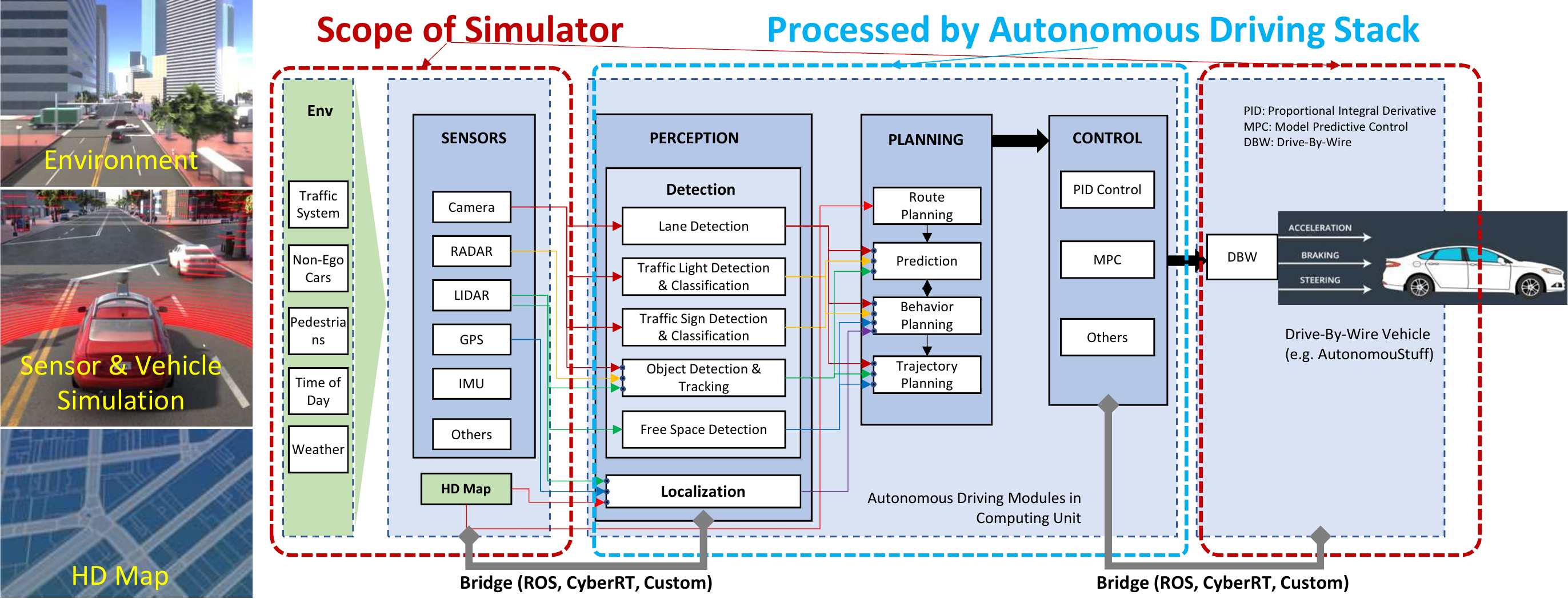}
\caption{High-level architecture of autonomous driving system and the roles of the simulation engine}
\label{Fig_SimulationEngine}
\end{figure*}

If the user's AD stack uses a custom runtime framework, a custom communication bridge interface can be easily added as a plug-in. Furthermore, LGSVL Simulator supports multiple AD systems connected simultaneously. Each AD system can communicate with the simulator through a dedicated bridge, enabling interaction between different autonomous systems in a unified simulation environment.

\subsection{Simulation Engine}

LGSVL Simulator utilizes Unity's game engine for simulation and takes advantage of the latest technologies in Unity, such as High Definition Render Pipeline (HDRP), in order to simulate photo-realistic virtual environments that match the real world.

Functions of the simulation engine can be broken down into: environment simulation, sensor simulation, and vehicle dynamics and control simulation of an ego vehicle. Fig.~\ref{Fig_SimulationEngine} shows the relationship between the simulation engine and the AD stack.

Environment simulation includes traffic simulation as well as physical environment simulation like weather and time-of-day. These aspects are important components for test scenario simulation. All aspects of environment simulation can be controlled via the Python API. 

The simulation engine of LGSVL Simulator is developed as an open source project. The source code is available publicly on GitHub, and the executable can be downloaded and used for free. 

\subsection{Sensor and Vehicle Models}

The ego vehicle sensor arrangement in the LGSVL Simulator is fully customizable. The simulator's web user interface accepts sensor configurations as JSON formatted text allowing easy setup of sensors' intrinsic and extrinsic parameters. Each sensor entry describes the sensor type, placement, publishing rate, topic name, and reference frame of the measurements. Some sensors may also have additional fields to further define specifications; for example, each LiDAR sensor's beam count is also configurable.

\begin{figure}[tb]
\centering
    \includegraphics[width=0.495\linewidth]{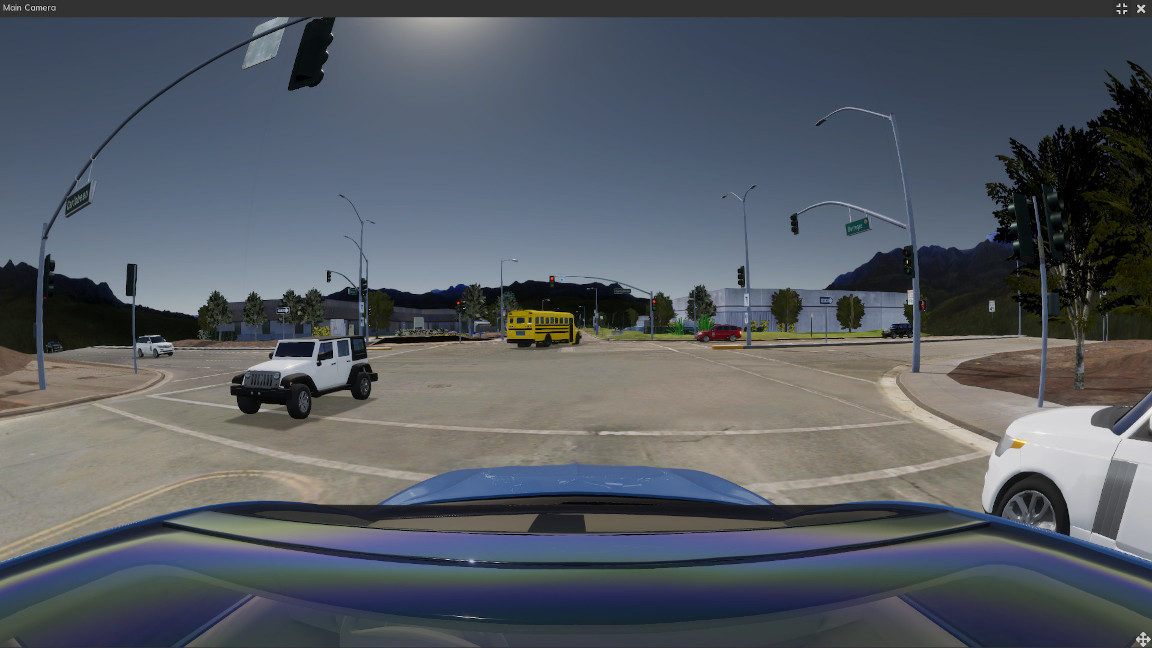}\hfill
    \includegraphics[width=0.495\linewidth]{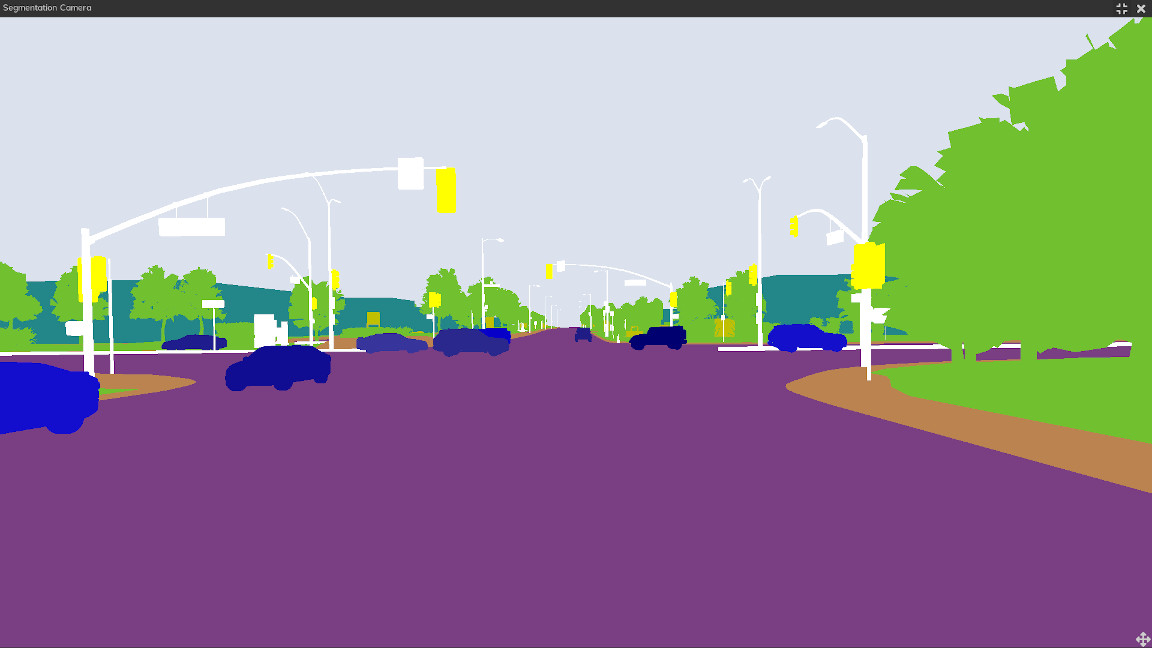}\\
    \vspace{.6mm}
    \includegraphics[width=0.495\linewidth]{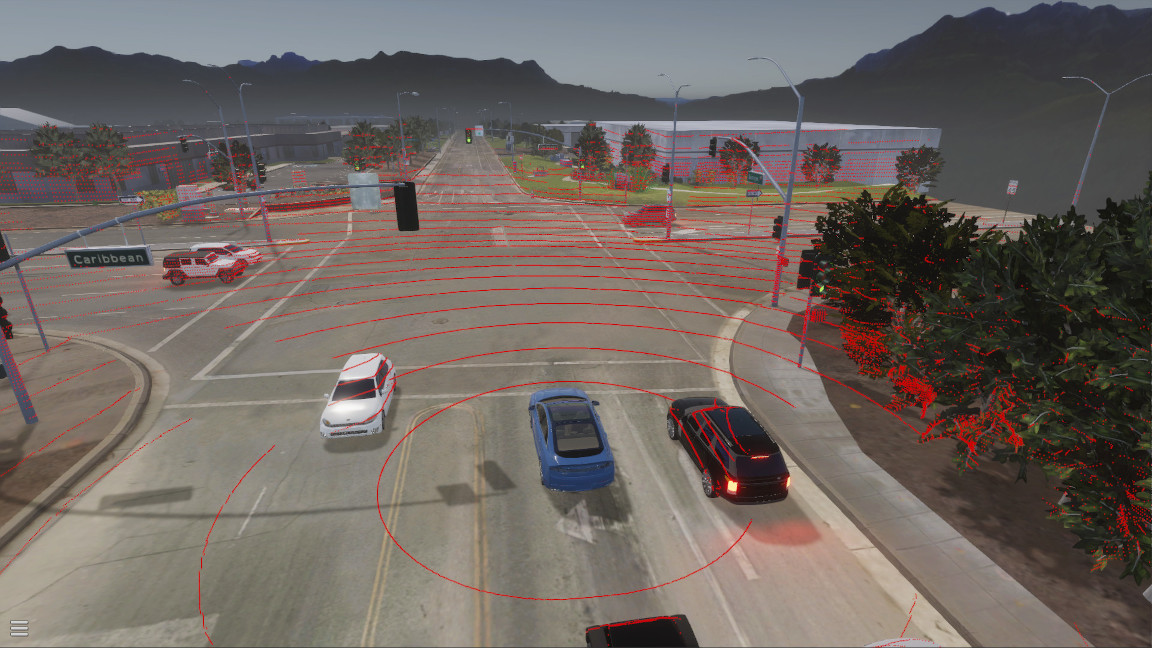}\hfill
    \includegraphics[width=0.495\linewidth]{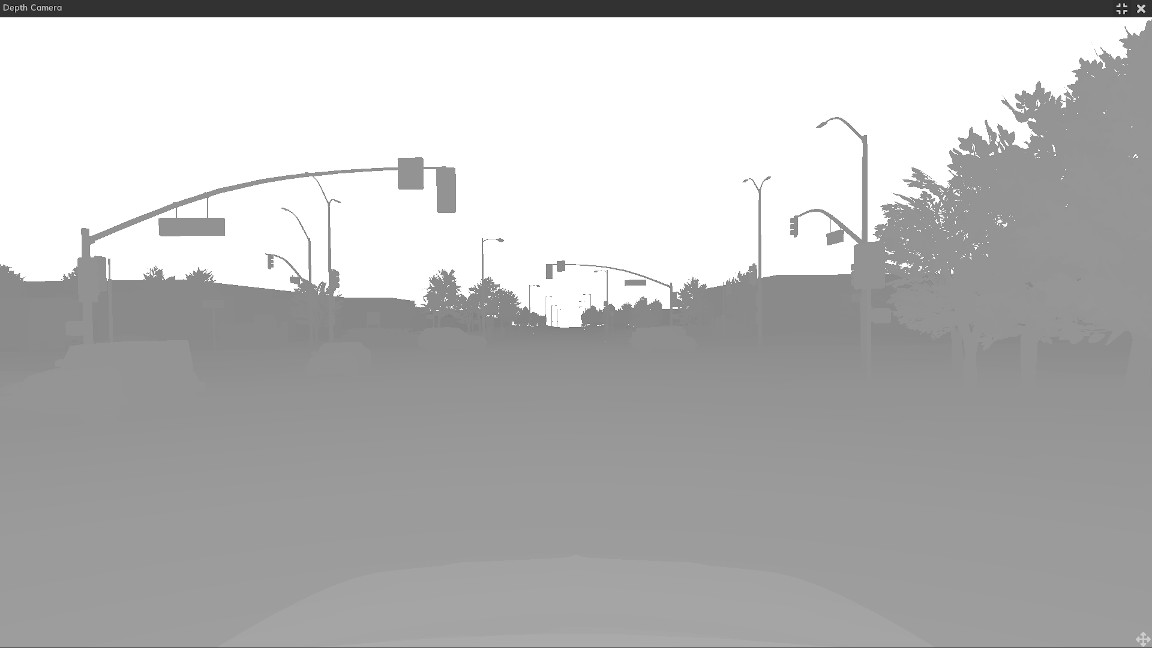}\\
    \vspace{.6mm}
    \includegraphics[width=0.495\linewidth]{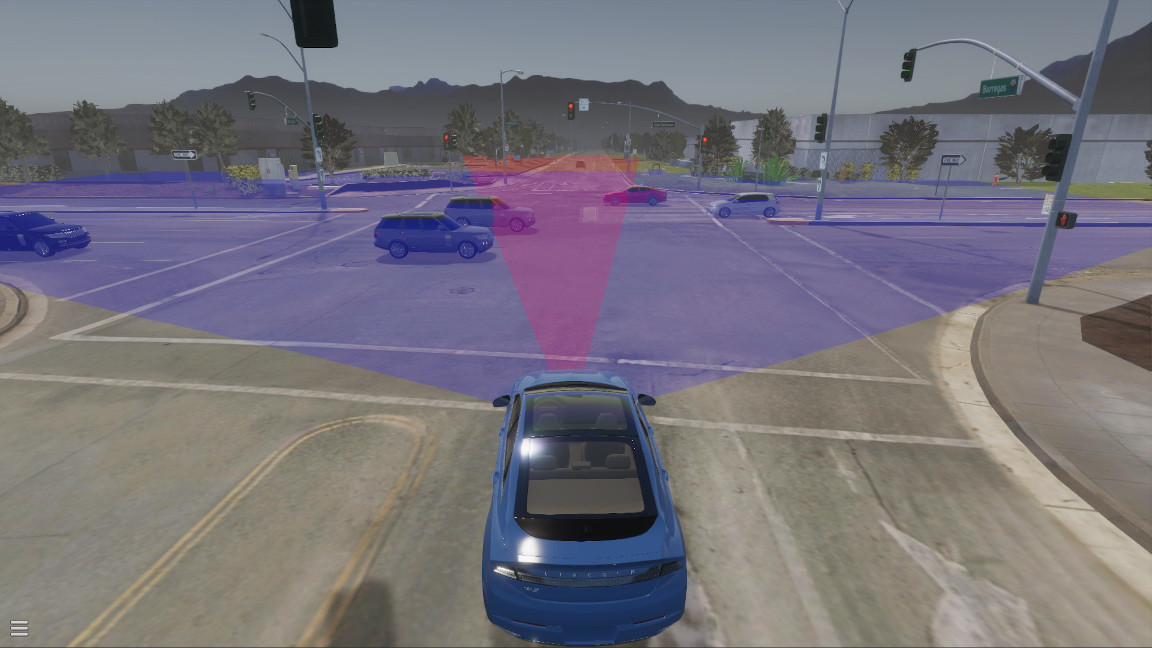}\hfill
    \includegraphics[width=0.495\linewidth]{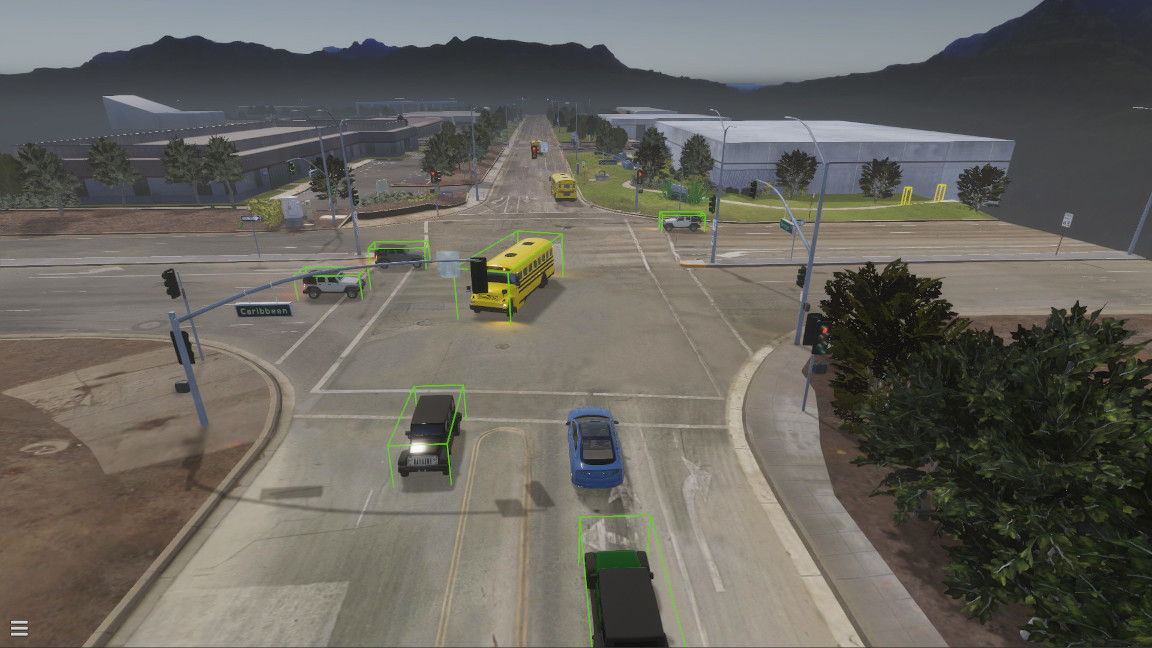}
\caption{Different types of sensors. Left (top to bottom): Fish-eye camera, LiDAR, Radar; Right (top to bottom): Segmentation, Depth, 3D Bounding Box.}
\label{Fig_Sensors}
\end{figure}

The simulator has a default set of sensors to choose from which currently include camera, LiDAR, Radar, GPS, and IMU as well as different virtual ground truth sensors. Users can also build their own custom sensors and add them to the simulator as sensor plugins. Fig.~\ref{Fig_Sensors} illustrates some of sensors in LGSVL Simulator: left column shows some physical sensors including fish-eye camera sensor, LiDAR sensor, and Radar sensor; right column shows some virtual ground truth sensors including segmentation sensor, depth sensor, and 3D bounding box sensor. 

For segmentation sensor, we combine semantic segmentation and instance segmentation. Users can configure which semantics get instance segmentation -- each instance of objects with these semantics will get different segmentation colors, and instances of other types of objects only get one segmentation color per semantic. For example, if the user configured only ``car'' and ``pedestrian'' to have instance segmentation, all buildings will have the same segmentation color, and all roads will have another segmentation color. Each car and each pedestrian will have different segmentation color, but all cars' color will be similar (e.g. all bluish) and same as pedestrians (e.g. all reddish). 

In addition to the default reference sensors, real world sensor models used in autonomous vehicle systems are also supported in LGSVL Simulator. These sensor plugins have parameters that match their real world counterparts, e.g. Velodyne VLP-16 LiDAR, and generate point clouds in the same format as real sensors. Furthermore, users can create their own sensor plugins to implement new variations and even new types of sensors not supported by default in LGSVL Simulator.

LGSVL Simulator provides a basic vehicle dynamics model for the ego vehicle. Additionally, the vehicle dynamics system is set up to allow integration of external third party dynamics models through a Functional Mockup Interface (FMI) \cite{FMI}, shared libraries that can be loaded into the simulator, or separate IPC (Inter-Process Communication) interfaces for co-simulation. As a result, users can couple LGSVL Simulator together with third party vehicle dynamics simulation tools to take advantage of both systems.

\subsection{3D Environment and HD Maps}

The virtual environment is an important component in autonomous driving simulation that enables providing many inputs to an AD system.

As the source of inputs to all sensors, the environment affects an AD system's perception, prediction, and tracking modules.
The environment affects vehicle dynamics which is the key factor for the vehicle control module. It also influences the localization and planning modules through changes to the HD map, which depends on the actual 3D environment. Finally, the 3D environment is the basis for environmental simulation including weather, time of day, traffic agents, and other dynamic objects.

\begin{figure}[tb]
\centering
    \includegraphics[width=.9\linewidth]{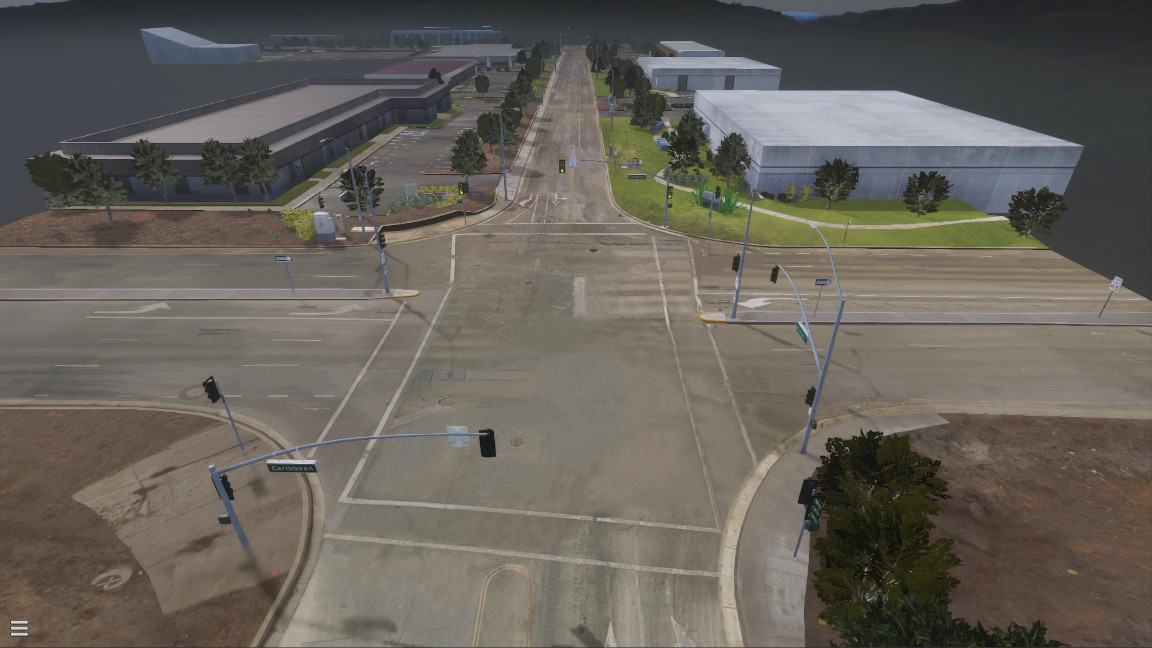}
\caption{Digital twin of Borregas Avenue.}
\label{Fig_BorregasAve}
\end{figure}

While synthetic 3D environments can be created and used in simulation, we can also replicate and simulate real world locations by creating a digital twin of a real scene from logged data (images, point cloud, etc.). Fig.~\ref{Fig_BorregasAve} shows a digital twin simulation environment we created for Borregas Avenue in Sunnyvale, California. In addition, we have collaborated with AAA Northern California, Nevada \& Utah to make a digital twin of a portion of GoMentum Station \cite{GoMentum}. GoMentum is an AV test facility located in Concord, CA featuring 19 miles of roadways, 48 intersections, and 8 distinct testing zones over 2,100 acres. Using the GoMentum digital twin environment, we tested scenarios in both simulation and with a real test vehicle at the test facility.

\begin{figure}[tb]
\centering
    \includegraphics[width=.9\linewidth]{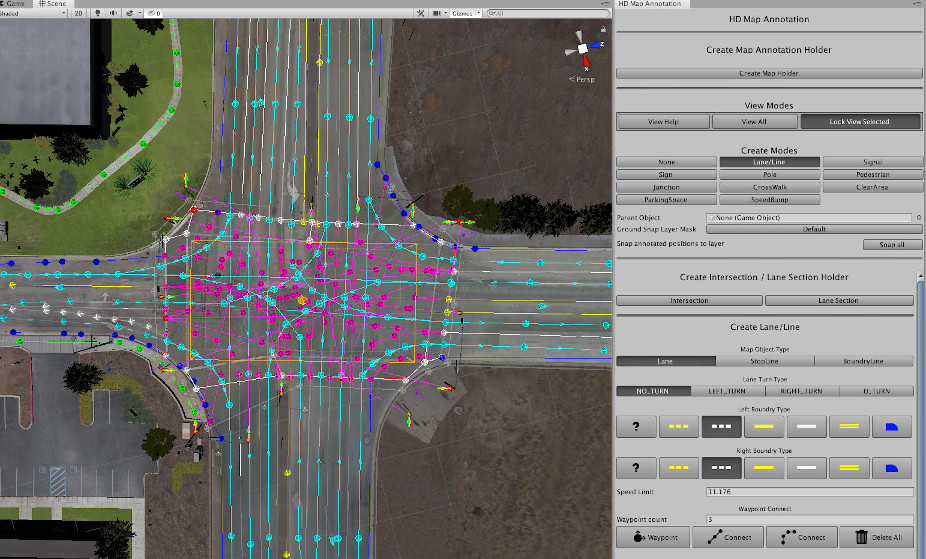}
\caption{HD Map example and annotation tool in LGSVL Simulator.}
\label{Fig_hdmap}
\end{figure}

LGSVL Simulator supports creating, editing, and exporting HD Maps of existing 3D environments. This feature allows users to create and edit custom HD map annotations in a 3D environment. While a 3D environment is useful as realistic simulation of the road, buildings, dynamic agents, and environment conditions which can be perceived and reacted on, map annotations can then be used by other agents in the environment that are part of a scenario (non-ego vehicles, pedestrians, controllable plugin objects). This means that vehicle agents in simulation will be able to follow traffic rules, such as traffic lights, stop signs, lanes, and turns, pedestrian agents can follow a annotated route, etc. 
As shown in Fig.~\ref{Fig_hdmap}, LGSVL Simulator HD map annotations have very rich information like traffic lanes, lane boundary lines, traffic signals, traffic signs, pedestrian walking routes, etc. On the right side of the figure, a user can make different annotations by choosing corresponding options under \emph{Create Modes}.

The HD map annotations can be exported into one of the several formats: Apollo 5.0 HD Map, Autoware Vector Map, Lanelet2, and OpenDrive 1.4, so users can use the map files for their own autonomous driving stacks. 
On the other hand, if a user has a real-world HD map in supported format, he/she can import the map into a 3D environment in LGSVL Simulator. The user will get the corresponding map annotations which are necessary for agents like vehicles, pedestrians to work. 
Currently, the supported HD map formats which can be imported are Apollo 5.0, Lanelet2, and OpenDrive 1.4.
With the ability to both import and export map annotations, a user could import HD maps sourced elsewhere, edit annotations, then export again to make sure that the HD maps used in LGSVL Simulator are coincident with that used by the user’s autonomous driving system.

\subsection{Test Scenarios}

Test scenarios consist of simulating an environment and situation in which an autonomous driving stack can be placed to verify correct and expected behavior. Lots of variables are included, such as time of day, weather, road condition, as well as distribution and movement of moving agents, e.g. cars, pedestrians, etc.

LGSVL Simulator provides a Python API to enable users to control and interact with simulated environments. Users can write scripts to create scenarios for their needs -- spawning and controlling NPC vehicles and pedestrians and set the environment parameters. With deterministic physics, scripting allows for repeatable testing in simulation. Improvements are continuously made to the platform to support better smart agents and traffic modeling to recreate scenarios that are as close to reality as possible.

We also collaborated with UC Berkeley using SCENIC \cite{FDG+2019} to generate and test thousands of different scenario test cases by randomizing various parameters. Results from testing those generated scenarios in simulation (using the GoMentum digital twin) then informed which scenarios and parameters would be most useful to test in the real world test facility.

\section{APPLICATIONS}
\label{Sec_Applications}

LGSVL Simulator enables various simulation applications for autonomous driving and more. Some examples are listed in this section. Since the ecosystem of LGSVL Simulator is an open environment, we believe users will extend this spectrum into more different domains.

\subsection{SIL and HIL Testing}

The LGSVL Simulator supports both software in the loop (SIL) and hardware in the loop (HIL) testing of AD stacks. 

For SIL testing, LGSVL Simulator generates data for different perception sensors, e.g. images for camera sensors and point cloud data for LiDAR sensors, as well as GPS and IMU telemetry data which are used by the perception and localization modules of an AD stack. This enables end-to-end testing of the users' AD stack. Furthermore, LGSVL Simulator also generates input for other AD stack modules to enable single module (unit) tests. For example, 3D bounding boxes can be generated to simulate output from a perception module as input for a planning module, so users can bypass the perception module (i.e. assuming perfect perception) to test just the planning module.

LGSVL Simulator supports a set of chassis commands, so that a machine running LGSVL Simulator can communicate with another machine running an AD stack which can then control the simulated ego vehicle using these chassis commands. This enables HIL testing where the AD stack is not able to differentiate inputs coming from a real car from simulation data and can send control commands to LGSVL Simulator in the same way it sends to the real car.

To verify the effectiveness of our simulation, we collaborated with UC-Berkeley to test thousands of scenarios generated by Scenic \cite{FDG+2019} in the digital twin of GoMentum Station \cite{GoMentum} and selected several representative scenarios to test in real GoMentum Station. The comparison of simulation results and real runs indicated the simulation is effective at identifying relevant tests for track testing with a real autonomous vehicle. More details of comparison can be found in \cite{FKV+2020}.

\subsection{Machine Learning and Synthetic Data Generation}

The LGSVL Simulator provides an easy-to-use Python API that enables collecting and storing camera images and LiDAR data with various ground truth information -- occlusion, truncation, 2D bounding box, 3D bounding box, semantic and instance segmentation, etc. Users can write Python scripts to configure sensor intrinsic and extrinsic parameters and generate labeled data in their own format for perception training. An example Python script to generate data in the KITTI format is provided on GitHub\footnote{https://www.lgsvlsimulator.com/docs/api-example-descriptions/\#collecting-data-in-kitti-format}.

Reinforcement learning is an active area of research for autonomous vehicles and robotics, often with the goal of training agents for planning and control. In reinforcement learning, an agent takes actions in an environment based on a policy, often implemented as a DNN, and receives a reward as feedback from the environment which in-turn is used to revise the policy. This process generally needs to be repeated through a large number of episodes before an optimal solution is achieved. The LGSVL Simulator provides out-of-the-box integration with OpenAI Gym \cite{OpenAIGym} through the Python API\footnote{https://www.lgsvlsimulator.com/docs/openai-gym/}, enabling the LGSVL Simulator as an environment that can be used for reinforcement learning with OpenAI Gym.

\subsection{V2X System}

In addition to sensing the world via equipped sensors, autonomous vehicles can also benefit from V2X (vehicle-to-everything) communications, such as getting information of other vehicles via V2V (vehicle-to-vehicle) and getting more environment information via V2I (vehicle-to-infrastructure). Testing V2X in real world is even more difficult than testing a single autonomous vehicle since it requires connected vehicles and infrastructure support. Researchers usually use simulator to test and verify V2X algorithms \cite{WWB+2019}. LGSVL Simulator supports creation of real or virtual sensor plug-ins which enables users to create special V2X sensors to get information from other vehicles (V2V), pedestrians (V2P), or surrounding infrastructures (V2I). Thus LGSVL Simulator can be used to test V2X systems as well as to generate synthetic data for training.

\subsection{Smart City}

Modern smart city systems utilize road-side sensors to monitor traffic flow. The results can be used to control traffic lights making traffic flow smoother. Such system requires different metrics to evaluate the traffic condition. One typical example is ``stop count'' -- the number of ``stop''s for a car to drive through an intersection, while a ``stop'' is defined as its speed falling down to lower than a given threshold for certain time. The ground truths of such metrics are difficult to be manually collected. LGSVL Simulator is also suitable for this kind of application. Using our sensor plug-in model, users can define a new type of sensor counting number of ``stop'' for a car since we have exact speed and location information. Our controllable plug-in allows users to customize traffic light and other special traffic signs which can be controlled via Python API.

\section{CONCLUSIONS}
\label{Sec_Conclusions}

We introduced LGSVL Simulator, a Unity-based high fidelity simulator for autonomous driving and other related systems. It has been integrated with Autoware and Apollo AD stacks for end-to-end tests, and can be easily extended for other similar AD systems. Several application examples are provided to show the capabilities of the LGSVL Simulator. 

The simulation engine is open source and the whole ecosystem is designed to be open, so that users can utilize LGSVL Simulator for different applications and add their own contributions to the ecosystem. The simulator will be continuously enhanced to address new requirements from the user community.

\section*{ACKNOWLEDGMENT}

This work is done within LG Electronics America R\&D Lab. We thank all the past and current colleagues who have contributed to this project. We also thank all external contributors on GitHub and all users who have provided feedback/suggestions to us.

\bibliographystyle{IEEEtran}
\bibliography{IEEEabrv,references}

\end{document}